\documentclass[acmtog,screen,nonacm]{acmart}

\usepackage{xspace}

\newcommand{\toolname}{GimmBO\xspace}%
\newcommand{\DesignSpace}{\mathcal{A}}
\usepackage{bm}

\usepackage{wrapfig}
\usepackage{graphicx}
\usepackage{multicol}
\usepackage{multirow}
\usepackage[dvipsnames]{xcolor}
\usepackage[table]{xcolor}
\usepackage{booktabs}
\usepackage{xspace}
\usepackage{amsmath}

\usepackage{amssymb}

\usepackage[mathletters]{ucs}
\usepackage{soul}
\usepackage[utf8]{inputenc}
\usepackage{bm}

\usepackage[ruled]{algorithm2e} %

\usepackage{layouts}
\makeatletter 
\newcommand{\layoutdetails}{%
\begin{tabular}{ll}
 \texttt{\textbackslash{textwidth}} & \printinunitsof{in}\prntlen{\textwidth} \\
\texttt{\textbackslash{linewidth}} & \printinunitsof{in}\prntlen{\linewidth} \\
Main text font &  \f@size pt \f@family \\
\sffamily \small Caption text font &  \sffamily \small \f@size pt \f@family \\
\end{tabular}%
}
\makeatother

\usepackage{dblfloatfix}
\usepackage{subcaption}

\usepackage{fontawesome5}

\providecommand{\A}{}
\providecommand{\B}{}

\providecommand{\I}{}

\providecommand{\W}{}

\providecommand{\x}{}

\providecommand{\balpha}{}

\renewcommand{\balpha}{\bm{\alpha}}

\renewcommand{\x}{\mathbf{x}}

\renewcommand{\A}{\mathbf{A}}
\renewcommand{\B}{\mathbf{B}}

\renewcommand{\I}{\mathbf{I}}

\renewcommand{\W}{\mathbf{W}}

\usepackage{booktabs} %

\usepackage[ruled]{algorithm2e} %

\SetAlFnt{\small}
\SetAlCapFnt{\small}
\SetAlCapNameFnt{\small}
\SetAlCapHSkip{0pt}

\begin{document}
\title[%
\toolname: Interactive Generative Image Model Merging via Bayesian Optimization%
]{\toolname: Interactive Generative Image Model Merging\\%
via Bayesian Optimization}

\author{Chenxi Liu}
\orcid{0000-0003-3613-1662}
\affiliation{%
  \institution{University of Toronto}
  \city{Toronto}
  \country{Canada}}
\email{liuchenxi0921@gmail.com}

\author{Selena Ling}
\orcid{0000-0001-6458-4488}
\affiliation{%
  \institution{University of Toronto}
  \city{Toronto}
  \country{Canada}}
\email{selena.ling@mail.utoronto.ca}

\author{Alec Jacobson}
\orcid{0000-0003-4603-7143}
\affiliation{%
  \institution{University of Toronto}
  \city{Toronto}
  \country{Canada}}
\affiliation{%
  \institution{Vector Institute}
  \city{Toronto}
  \country{Canada}}
\email{jacobson@cs.toronto.edu}

\begin{abstract}
Fine-tuning-based adaptation is widely used to customize diffusion-based image generation, leading to large collections of community-created adapters that capture diverse subjects and styles.
Adapters derived from the same base model can be merged linearly, enabling the synthesis of new visual results within a vast and continuous design space.
To explore this space, current workflows rely on manual slider-based tuning, an approach that scales poorly and makes merging coefficient selection difficult, even when the candidate set is limited to 20–30 adapters.
We propose \toolname to support interactive exploration of adapter merging for image generation through Preferential Bayesian Optimization (PBO).
Motivated by observations from real-world usage, including sparsity and constrained coefficient ranges, we introduce a two-stage BO backend that improves sampling efficiency and convergence in high-dimensional spaces.
We evaluate our approach with simulated users and a user study, demonstrating improved convergence, high success rates, and consistent gains over BO and line-search baselines, and further show the flexibility of the framework through several extensions.
\end{abstract}



\begin{teaserfigure}
  \includegraphics[width=\linewidth]{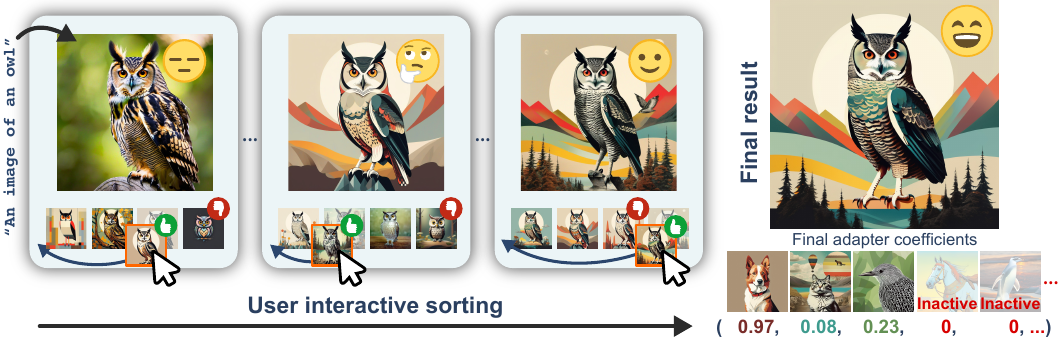}
  \caption{
  Given a prompt, default image generation may appear reasonable yet fail to match a user’s creative intent (left).
  \toolname enables users to explore weighted combinations of customization \emph{adapters} (30 in this example) through preference feedback, producing an image that better reflects the desired style (right, corresponding adapter coefficients below).
  \faSearchPlus\ Readers are encouraged to zoom in on images for finer details throughout the paper.
  }
  \label{fig:teaser}
\end{teaserfigure}

\maketitle

\section{Introduction}
\label{sec:intro}

\begin{figure}[t]
  \includegraphics[width=\linewidth]{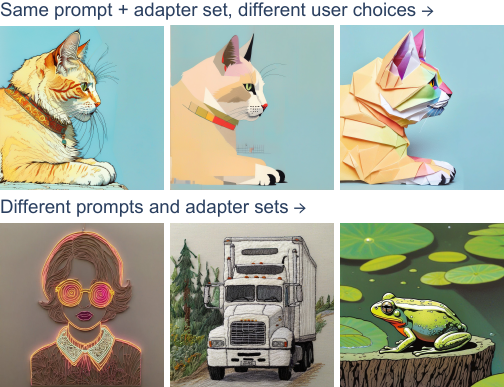}
  \caption{
  The adapter design space is vast.
  \toolname assists users to produce substantially different yet equally satisfying results.
  }
  \label{fig:intro}
\end{figure}

Fine-tuning-based customization of diffusion models is widely used to synthesize subjects or styles underrepresented or missing from the original training data~\cite{ruiz2023dreambooth,hu2022lora,zheng2024stylebreeder}.
These fine-tuned models, or \emph{adapters}, are lightweight to train, store, and share, leading to large community-curated adapter libraries.
Adapters derived from a common base model can be merged to combine visual characteristics, forming a vast and continuous design space (Figs.~\ref{fig:teaser}, \ref{fig:intro}), and are commonly used in multi-adapter form on image generation platforms  (69\% of adapter-based generations on  
\begin{wrapfigure}{r}{0.44\linewidth}
  \includegraphics[width=\linewidth,trim=1.2cm 1.cm 0.0cm 0.65cm]{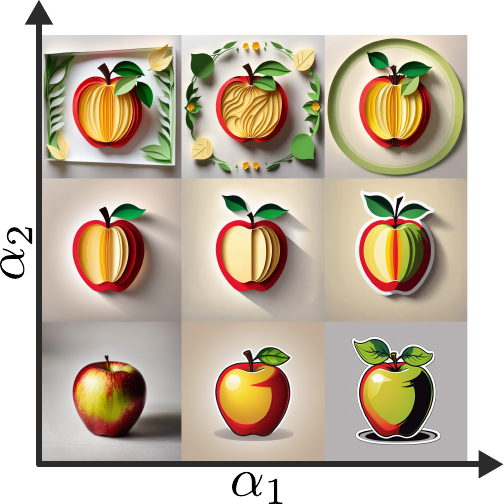}
\end{wrapfigure}
Civitai\footnote{A community platform for image generation. Statistics collected in December~2025.}).
In practice, users explore this design space by manually tuning per-adapter slider values (e.g., $\alpha$s in inset); however, this approach scales poorly (with steep degradation beyond 5-10 sliders~\cite{dang2022ganslider}) and makes coefficient selection difficult even after users limit the candidate set to 20–30 relevant adapters.
Existing interactive design optimization systems are largely confined to low-dimensional design spaces (3-12D) in real applications~\cite{koyama2017sequential,koyama2020sequential,koyama2022bo,niwa2025cooperative,mo2024cooperative}, making 20-30D high-dimensional in this context.

We propose \toolname to allow users to explore and identify suitable coefficient combinations over 20–30 adapters during interactive design.
This interactive human-in-the-loop setting raises two key challenges: user preferences define a black-box objective, and each evaluation is costly due to model inference and user interaction.
We address these challenges using human-in-the-loop Bayesian optimization (BO), a sample-efficient approach commonly used in design tasks (Sec.~\ref{sec:related}).
However, this setting lies outside the effective range of existing BO-based interactive design methods due to its high dimensionality.
Surveying real-world practice, we observe that effective adapter merges are typically \emph{sparse}, involving only a small subset of adapters, and that merging coefficients must lie within a \emph{limited magnitude range} to avoid degraded or incoherent generations (Sec.~\ref{sec:how}).
Motivated by these observations, we introduce a two-stage BO backend that first operates in a conservative search space for efficient sampling and then exploits sparsity and early variable selection to accelerate convergence (Sec.~\ref{sec:method}).

We extensively evaluate our approach through quantitative experiments with both simulated and real users.
Our primary experiments operate in a 20-dimensional design space of stylization adapters.
Using simulated users on test sets constructed from realistic input distributions and real adapter collections, we demonstrate the practical usefulness of \toolname under controlled and repeatable conditions.
We then conduct a user study to validate effective support for human preference-driven interaction.
Scalability is further assessed through stress tests with simulated users in 30D and 40D spaces.
Across all evaluations, our approach exhibits faster convergence and higher-quality results, highlighting the benefits of incorporating domain-specific structure into the BO backend.
Finally, we explore several extensions, including merging beyond style to incorporate content, as well as downstream applications.
We also show that \toolname integrates naturally with community model retrieval approaches~\cite{luo2024stylus}, enabling an end-to-end workflow that automatically identifies a candidate set of 20–30 adapters and supports interactive exploration of their combinations.
Our implementation and supplementary materials are available at \url{https://gimmbo-project.github.io}.

\section{Related Work}
\label{sec:related}

\paragraph{Customized image generation}
Diffusion model customization\linebreak adapts pretrained models to specific subjects and styles, enabling personalized image generation beyond the original training distribution.
Early approaches bind new concepts via learned textual tokens, as in textual inversion~\cite{gal2022image}, with extensions using multi-token or time-dependent embeddings~\cite{voynov2023p+,alaluf2023neural}.
Complementary fine-tuning methods adapt model parameters from few reference images, including DreamBooth~\cite{ruiz2023dreambooth} and selective parameter updates to reduce overfitting and storage~\cite{kumari2023multi}.
To improve efficiency, parameter-efficient fine-tuning introduces lightweight trainable components such as style-specific modules~\cite{sohn2023styledrop} and low-rank adaptation (LoRA)~\cite{hu2022lora}, now widely adopted for diffusion model customization.
Beyond per-subject training, pre-trained adaptation methods enable fast one-shot personalization by conditioning on reference inputs~\cite{li2023blip,ye2023ip}.
See \citet{zhang2025survey} for a comprehensive survey.

\paragraph{Model merging}
In this setting, multiple adapters fine-tuned from a shared base model are combined to produce a merged effect~\cite{yadav2024survey}.
In the text-to-image domain, merging has been used to compose multiple concepts~\cite{gu2023mix,po2024orthogonal,dalva2025lorashop} or to combine content with style~\cite{shah2024ziplora,frenkel2024implicit,ouyang2025k}.
Recent works have also explored, in addition to adapter selection, generating the entire image-generation workflow~\cite{gal2025comfygen,gadot2025policy}.
In contrast to these settings, which treat merging as a discrete operation or focus on fixed combinations, we focus on treating adapter merging as a continuous design problem, where the \emph{strength} of each component can be adjusted 
(e.g., ``+\texttt{[Adapter]}'' vs.\ ``+0.5\ \texttt{[Adapter]}'').
Our quantitative evaluation focuses on merging multiple style adapters (Sec.~\ref{sec:quant}), reflecting common real-world usage and a setting that is comparatively less structured and under-explored than style-content composition.
More broadly, \toolname generalizes naturally to other forms of adapter composition, including subject- and content-based merging as demonstrated in Sec.~\ref{sec:qual}.
We adopt linear merging, a simple yet effective method that is widely used in practice, while remaining compatible with more advanced merging techniques.
A related line of work studies community-model retrieval~\cite{liu2024loraworththousandpictures,luo2024stylus,sonmezer2025loraverse}, which selects and combines adapters shared online, e.g., via LLM-based textual retrieval.
These approaches are complementary to ours, as they can help narrow the candidate set, but they do not explicitly address continuous control over merging weights; we demonstrate how \toolname integrates naturally with such retrieval methods in Sec.~\ref{sec:qual}.

\begin{figure}[t]
  \includegraphics[width=\linewidth]{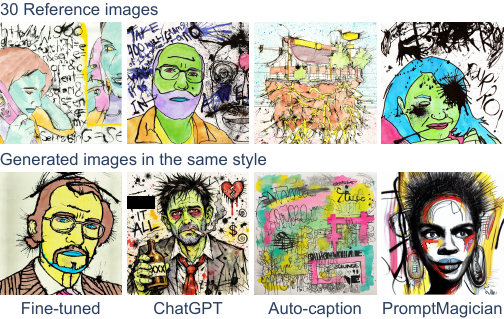}
  \caption{
  Given 30 reference images by an artist with few publicly available works (top), we attempt to reproduce the style (bottom). Fine-tuning an SDXL LoRA adapter closely matches the target style, while prompt-based approaches, including ChatGPT image creation (a dark block indicates a censored word), auto-captioning followed by SDXL generation, and PromptMagician~\cite{feng2023promptmagician}, achieve only partial similarity, highlighting the limitations of text-based elicitation for underrepresented styles.
  }
  \label{fig:prompt}
\end{figure}

\paragraph{Prompt-based customization}
Existing methods explore customized image generation by adapting text prompts based on human interaction, often through visualization interfaces~\cite{brade2023promptify,feng2023promptmagician,guo2024prompthis,adamkiewicz2025promptmap} or conversational agents~\cite{wen2026adaptive}.
Such prompt-based approaches can be applied to scenarios like style mixing by iteratively refining textual descriptions.
However, they implicitly assume that the desired styles are already encoded in the base model and can be elicited through prompting.
As a result, they often struggle to capture out-of-distribution or highly specific styles compared to adapter-based approaches.
To illustrate this limitation, we consider reproducing the style of an artist with only a small number of publicly available works, which are likely underrepresented in the base model’s training distribution (Fig.~\ref{fig:prompt}).
In an adapter-based stylization setting (Stable Diffusion XL with LoRA adapters), fine-tuning on 30 images was sufficient to faithfully capture the target style.
In contrast, prompt-based approaches using ChatGPT, either directly from reference images or via auto-captioning followed by SDXL generation, produced results with only partial similarity.
We also evaluated PromptMagician~\cite{feng2023promptmagician}, a visualization-based prompt exploration tool, but even careful prompt search did not achieve comparable results, suggesting that certain stylistic nuances are difficult to express purely through text.

\begin{figure}[t]{\textbf{}}
  \includegraphics[width=\linewidth]{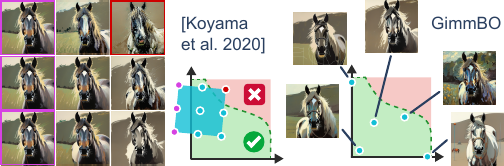}
  \caption{
    Previous BO-based interaction methods~\cite{koyama2020sequential}, which do not account for the structure of the adapter-merging design space, may propose clamped or degraded samples in high-dimensional settings.
  }
  \label{fig:related}
\end{figure}

\paragraph{Human-in-the-loop design optimization}
Many design tasks are inherently subjective, making human judgment essential to the design process and typically expressed through iterative choices over design parameters.
However, as the number of parameters grows, this exploration becomes difficult to perform manually, even with as few as 5–10 parameters~\cite{dang2022ganslider}.
Researchers have explored human-in-the-loop optimization to support efficient exploration of complex design spaces.
Prior work includes interaction paradigms such as slider interfaces~\cite{desai2019geppetto,koyama2014crowd,shimizu2020design,shugrina2015fab} and design galleries~\cite{marks97,ngan2006image,shapira2009image} to reduce direct parameter manipulation, with Bayesian Optimization (BO) later adopted as a sample-efficient backend within these settings for visual design tasks~\cite{koyama2017sequential,koyama2020sequential}.
BO has also been adapted to specific domains such as animation~\cite{brochu2010bayesian}, melody~\cite{zhou2020generative}, font~\cite{kadner2021adaptifont,tatsukawa2025fontcraft}, and game levels~\cite{khajah2016designing}.
Our method falls within this line of domain-specific BO approaches, and we explicitly compare against previous BO-based interaction methods for general visual design tasks~\cite{koyama2017sequential,koyama2020sequential} (Sec.~\ref{sec:quant}).
While effective in lower-dimensional settings, these previous methods are ill suited to our 20–30D adapter-merging scenario:~\citet{koyama2017sequential} relies on single-direction updates that do not scale, and~\citet{koyama2020sequential} assumes interior solutions and ignores poor-quality regions, leading respectively to duplicated clamped samples and degraded configurations (Fig.~\ref{fig:related}, Sec.~\ref{sec:how}).
Recently, researchers explore how to make the design process more cooperative~\cite{koyama2022bo,mo2024cooperative,niwa2025cooperative}.
These methods emphasize interface design, whereas we contribute an optimization backend tailored to the model-merging problem.
The two are complementary, and integrating them would be an interesting future direction.

Most existing mechanisms for dealing with high-dimensionality in BO are difficult to apply directly in our setting: embedding methods~\cite{wang2016bayesian,raponi2020high,chiu2020human} and trust regions~\cite{diouane2023trego} do not account for solution sparsity without adaptation, and our human-preference objective is unlikely to be additive.
We therefore apply BO solvers with built-in variable-selection capabilities~\cite{eriksson2021high,liu2023sparse}, which are better matched to our problem.

\section{How do people merge adapters today?}
\label{sec:how}
Given a generative image \emph{base model} with weights $\W_0$, fine-tuned adapters work by
updating the weights ($\W_0 + \Delta \W$) via further loss optimization on a new, specialized dataset.
For example, an adapter may be created by fine-tuning the SDXL base model to generate vector-art-style images.
A different SDXL-based adapter may be fine-tuned to generate images of papercraft artwork.
Sites like Civitai allow users to not only pick the strength of a given adapter but also blend multiple adapters linearly to combine their effects:
\begin{equation}
\label{equ:merge}
\W_\text{merged} = \W_0 + \sum_{i=1}^n \alpha_i \Delta \W_i,
\end{equation}
where $\alpha_i$ is the merging coefficient controlling the strength of the $i$th adapter's weight update $\Delta \W_i$.
It is both well known in practice and well supported by linear mode connectivity theory \cite{ilharco2023editing,frankle2020linear} that this simple linear combination of adapters of the same base model can result in meaningful combinations of individually learned abilities (e.g., image styles).
In practice, most adapters are parameter-efficient fine-tunings such as LoRAs \cite{hu2022lora} (where weight updates are the product of two low rank matrices $\Delta \W = \B \A$), but this is not a requirement.

The combination of image styles is complex to describe, hard to predict, and exciting. On sites like Civitiai, users experiment with different selections of adapters and different merging coefficients to create and explore new styles that would otherwise be hard to achieve through prompt conditioning of the base model or with an individual existing adapter.
The choice of merging coefficients $\alpha_i$ is non-trivial. People typically rely on trial and error: pick new $\alpha_i$ values with a slider or numeric entry, wait for a generated image, then iterate on their choice.
Despite this painful process, there exist enough examples of successfully merged adapters to identify two common practices which guide our method's design.

\begin{wrapfigure}{r}{0.5\linewidth}
  \includegraphics[width=\linewidth,trim=1.5cm 1.5cm 0.0cm 1.5cm]{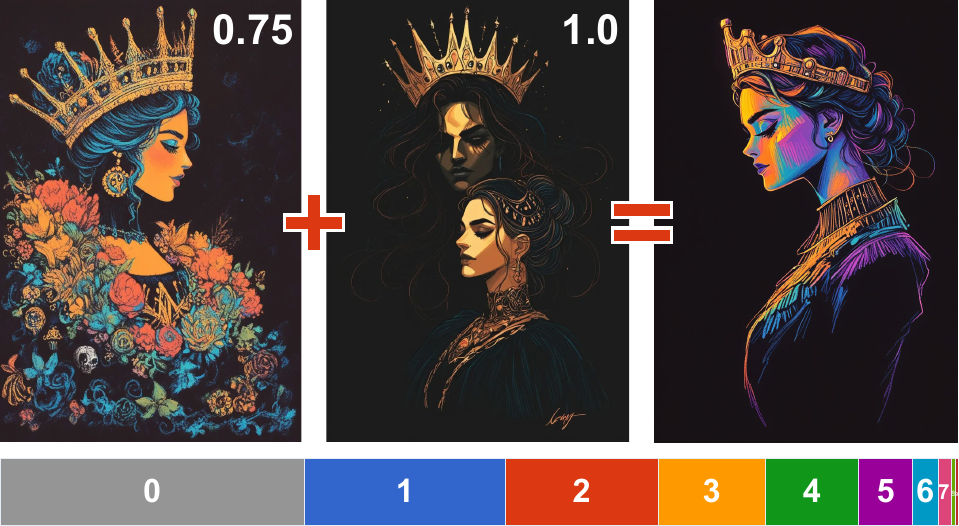}
\end{wrapfigure}
\paragraph{Sparsity}
We collected the generated images hosted on Civitai and analyzed their provenance.
Of the 20,780 images with sufficient meta data, we found that a strong majority are created using at least one adapter (68\%) and most of those are created by merging between two to nine adapters (69\%) (see inset
example of a generation using a merge of two adapters and a 
bar graph of adapter counts across the collected images).
We speculate that this reflects both a preference for interpretability and avoidance of technical issues such as model interference.
We will leverage this sparsity to guide our optimization from an initial large and possibly crudely collected candidate set ($n=20-30$) to a small active set of adapters (those with non-zero merging coefficient $\alpha_i\neq 0$).
Despite the heavy-tailed distribution of contributions (out of 3,533 users, top 1\% contribute 33.3\% of works and top 5\% 57.8\%), statistics remain consistent after removing the top 1\% and 5\% of users, indicating no significant bias.

\paragraph{Coefficient magnitudes}
By construction, a successfully trained fine-tuned adapter works well (in isolation) with unit magnitude ($\alpha=1$). Linear mode connectivity supports that values within or near $[0,1]$ will also behave well, diminishing the learned effect as $\alpha$ approaches zero.
When merging $n$ adapters, not all combinations of
\begin{wrapfigure}{r}{0.45\linewidth}
    \includegraphics[width=\linewidth,trim=1.9cm 1.5cm 0.0cm 1.5cm]{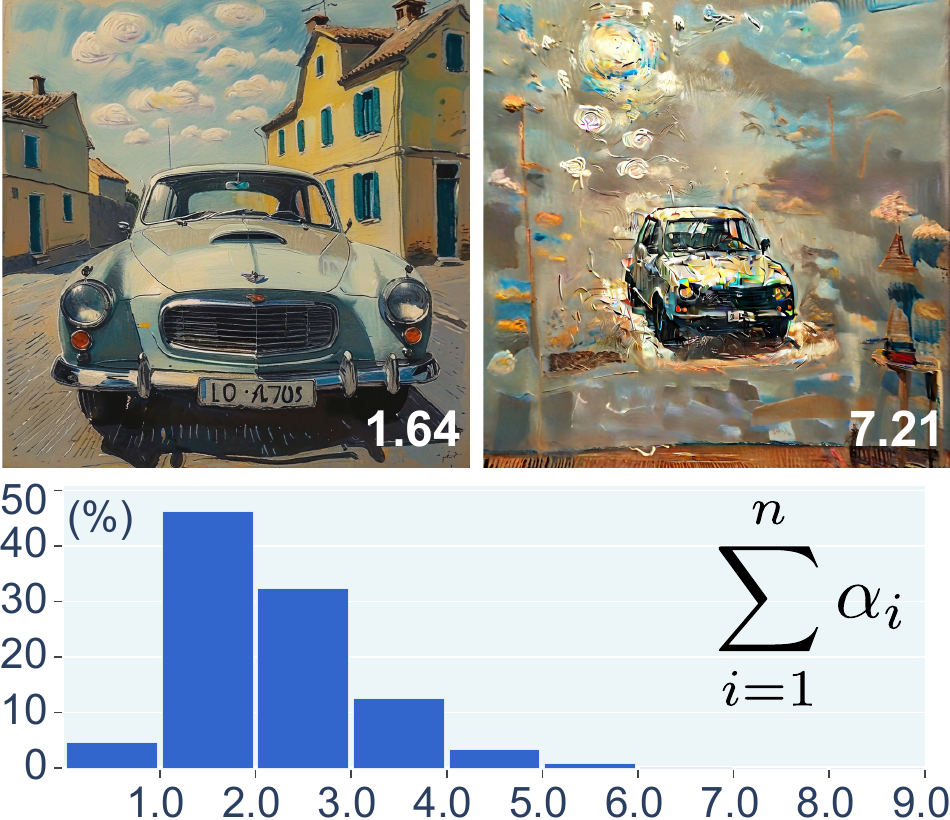}
\end{wrapfigure}
merging coefficients in the entire $[0,1]^n$ hypercube will result in meaningful generations.
For example, the inset images generated by merging $n=20$ adapters produces a meaningful stylized image for 
a small sum of coefficients but loses coherence for a large sum.
Analyzing the generated images from Civitai, we find that the sum of merging coefficients $\sum_{i=1}^n \alpha_i$ falls off quickly, with most sums falling between 1.0 and 3.0 (inset histogram).
We speculate that the true feasible set of merging coefficients is fuzzy, hard to precisely define and adapter dependent.
We nevertheless leverage our observation by modeling this feasible set with a coarse proxy geometry. We will show that enforcing coefficients to stay within this set during early stages and loosening the restriction later will improve convergence speed and final accuracy.

\section{Our \toolname Method}
\label{sec:method}
Given a candidate set of $n$ (typically $20\text{-}30$) adapters, at the end of our interactive design process, our goal is to output a vector of merging coefficients $\balpha = [\alpha_1, \dots, \alpha_n]$ which resulting in maximally satisfying image generations, as subjectively judged by the human user.
We conceptually model this as an optimization problem:
\begin{align}
    \mathop{\text{argmax}}_{\balpha \in \mathcal{\DesignSpace}} f\left(g\left(\balpha\right)\right),
\end{align}
where $\mathcal{A}$ is the merging coefficients design space, referred to as the \emph{search space} in our optimization (a yet-to-be-defined subregion of $[0,1]^n$), $g$ is the fixed function that maps a choice of merging coefficients to a generated image (eliding text prompts or other conditioning parameters), and $f$ models the user's assigned utility of a given image. Our quest to find the best $\balpha$ values naturally relies on simultaneously learning a good surrogate approximation of $f$ from limited user feedback on generated images so far. 
We propose using human-in-the-loop Bayesian optimization.

\begin{figure}[t]
  \includegraphics[width=\linewidth]{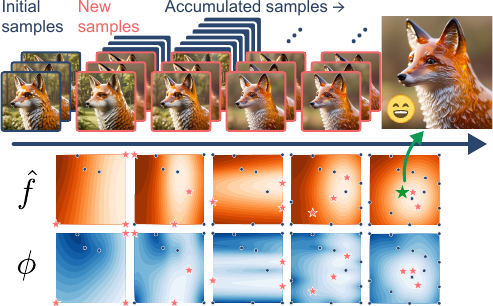}
  \caption{
    Starting from initial samples, \toolname iteratively proposes new sample batches (pink stars and images), guided by an acquisition function (blue) over a surrogate model $\hat{f}$ updated with accumulated samples (blue dots and images).
    In this 2D example, \toolname rapidly converges to the target.
  }
  \label{fig:method}
\end{figure}

\begin{figure*}[t]{\textbf{}}
  \includegraphics[width=\linewidth]{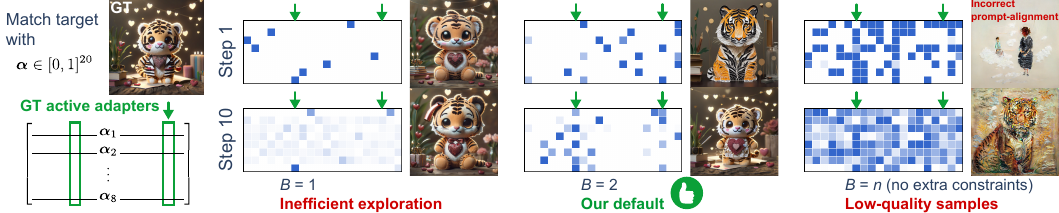}
  \caption{
  Batch samples at Steps~1 and~10 in a 20D matching task ($q=8$).
  Blue cells denote \textcolor[HTML]{3366CC}{\textbf{non-zero coefficients}} (darker = larger magnitude), white cells indicate zeros; green arrows mark ground-truth active axes.
  With identical initialization and random seed, $B=2$ performs best compared to $B=1$ and $B=n$.
    }
  \label{fig:pattern}
\end{figure*}

\toolname works iteratively (Fig.~\ref{fig:method}).
Each iteration, we ask the user to rank a small number of images generated by different merging coefficient vectors.
This ranking information is used to update: a prior over surrogates $\hat{f}$ that aim to proportionally approximate the unknown user preference function $f$; and an acquisition function 
whose (local) maxima represent informative points for reducing uncertainty that the maximizer of $\hat{f}$ matches that of $f$.
The acquisition function is designed so that sampling it balances \emph{exploitation} of regions predicted to yield high utility scores with \emph{exploration} of uncertain regions that improve the surrogate model.

\subsection{Preferential Surrogate}
We model the surrogate $\hat{f}$ as a latent function over merging coefficients $\balpha$ equipped with a Gaussian process prior, conditioned on observed data $\mathcal{D}$:
\begin{align}
    \hat{f}(\balpha \mid \mathcal{D} ) \sim \mathcal{GP}\left( \mu(\balpha), k(\balpha,\balpha')\right),
\end{align}
fully specified by a mean function $\mu$ and covariance function $k$.

Unlike classic Gaussian processes, we cannot expect users to provide data in the form of direct numeric utility values~\cite{brochu2010bayesian,tsukida2011analyze}.
Instead, we collect observed data in the form of $|\mathcal{D}|$ pairwise comparisons:
\begin{align}
\mathcal{D} = \left\{ (\balpha_i, \balpha_j) \;\middle|\; \balpha_i \succ \balpha_j \right\},
\end{align}
where $\balpha_i \succ \balpha_j$ encodes that the user preferred the generated image $g(\balpha_i)$ over $g(\balpha_j)$.
We assume transitivity of user rankings, so 
presenting the user with $N$ images and them to rank the top $k$ results in up to $k(k-1)/2 + k(N-k)$ new inequalities in $\mathcal{D}$ (all pairs in the top $k$ + each top $k$ versus the remaining).
Given any update to $\mathcal{D}$, 
$\hat{f}$ is relearned by first employing the preference learning approach of \citet{chu2005preference} 
to convert the pairwise inequalities into latent function values $\hat{f}(\balpha_i)$ for each observed $\balpha_i$. Then the posterior mean and covariance of $\hat{f}$
is fit to these values using a sparse axis-aligned subspace (SAAS) prior \cite{eriksson2021high}
(see supplemental for further details).

\subsection{Search space}
\label{sec:simplex}
We consider two choices for the search space of merging coefficients. The first is the entire unit hypercube:
\begin{align}
\mathcal{A} = [0,1]^n.
\end{align}
This choice works well when the size of the candidate set $n$ is small, but when $n$ is large sampling merging coefficient vectors $\balpha \in [0,1]^n$ often results in meaningless or poor quality images, wasting precious user feedback.

Instead, when the candidate set is large, we define the search space to model our observation that people tend to merge adapters with small total magnitude. In particular, we propose modeling the space with a \emph{$B$-capped simplex}:
\begin{align}
\mathcal{A} = \Delta_B = \left\{ \balpha \in [0,1]^n \ \middle|  \ \sum_{i=1}^n \alpha_i \leq B \right\}.
\end{align}
Relative to the volume of $[0,1]^n$, 
this geometry has rapidly diminishing volume as $n$ increases \cite{marichal2006slices}:
\begin{align}
\operatorname{Vol}(n;\Delta_B)
=
\frac{1}{n!}
\sum_{k=0}^{\lfloor B \rfloor}
(-1)^k \binom{n}{k} (B-k)^n.
\end{align}
For example, $\operatorname{Vol}(10;2) \approx 0.03\%$ and $\operatorname{Vol}(30;2) < 10^{-21}\%$.
We will show that choosing $\mathcal{A} = \Delta_B$ for early iterations works well, while relaxing $\mathcal{A}=[0,1]^n$ in later iterations allows polished accuracy.

\subsection{Initialization}
\label{sec:init}
With our surrogate in place, we initialize samples of merging coefficient values $\balpha$ from $\mathcal{A} = \Delta_B$, for which images are generated and shown to the user for ranking.
Due to the diminishing volume of $\Delta_B$ for large $n$, naive rejection sampling from $[0,1]^n$ would be intractably slow.
We propose to use a modified Dirichlet or ``stick breaking'' process. Given a uniformly random sample $\mathbf{x} \in [0,1]^n$, we sample $\balpha$ via an coordinate-dependent inductive process:
\begin{align}
    \alpha_i = \mathop{\text{min}}(x_i R_{i-1},1) \text{ and }
R_i = R_{i-1} - \alpha_i \text{ with } R_0 = B.
\end{align}
This distribution is known to exhibit coordinate-ordering bias, which we mitigate by picking a random coordinate ordering for each sample.
Finally, we further sparsify samples by zeroing entries below a threshold $\tau$.

\subsection{Acquisition}
We adopt an upper confidence bound (UCB) acquisition function \cite{srinivas2009ucb}, which provides a simple and interpretable hyperparameter $\lambda$ to balance exploration and exploitation.
First written as a function of a single $\balpha$ vector, UCB is a simple sum of mean and variance terms: \newcommand{\asingle}{\phi}
\begin{align}
    \asingle(\balpha \mid \hat{f}) = \mu(\balpha) + \lambda \sqrt{k(\balpha,\balpha)}.
\end{align}
Each user-facing iteration we sample $q$ merging coefficient vectors $\{\balpha_1,\dots,\balpha_q\}$ using a batched-version of $\asingle$ maximizing the expectation of maximal single acquisition value \cite{Wilson2017}:
\begin{align}
\mathop{\text{max}}_{\{\balpha_1,\dots,\balpha_q\}} & \quad
\mathbb{E}_{\asingle} \left[ 
\mathop{\text{max}}_{j=1,\dots,q} \asingle(\balpha_j \mid \hat{f})
\right] \\
    \text{subject to:}&\quad \balpha_j \in \mathcal{A} \text{ for } j=1, \dots, q.
\end{align}
When $\mathcal{A} = [0,1]^n$, we can easily randomly sample starting positions and run L-BFGS-B with multiple random restarts to ascend to a set of local maxima.

When the search space is the $B$-capped simplex $\mathcal{A} = \Delta_B$, we treat the stick-breaking process in the previous Section~\ref{sec:init} as a parameterization of feasible merging coefficients as a function of points in the unit hypercube $\balpha(\mathbf{x})$. We draw starting positions by sampling $\mathbf{x}\in [0,1]^n$ and multi-restart L-BFGS-B ascent treating $\mathbf{x}$ as optimization parameters.
Fig.~\ref{fig:pattern} demonstrates the impact of different choices of $B$, and we set $B=2$ by default.

Images are generated for each of the newly acquired $\balpha_j$ samples.
These images, combined with those of the current top-ranked merge coefficient vector and a few previously seen samples, are presented to the user for the next round of ranking.

\subsection{Two-stage Optimization}
\begin{wrapfigure}{r}{0.2\linewidth}
\includegraphics[width=\linewidth,trim=3.2cm 2.3cm 0.0cm 2.7cm]{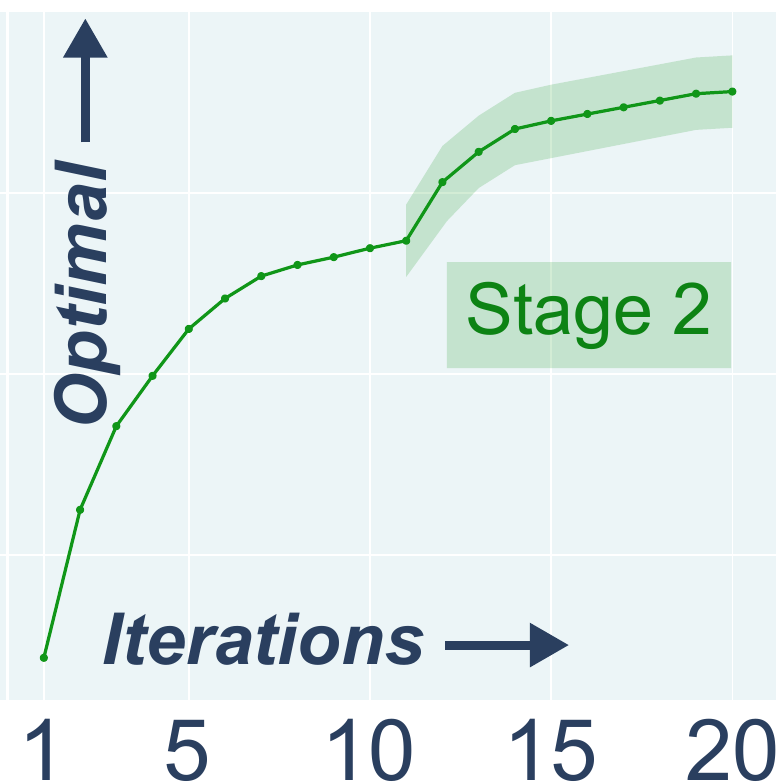}
\end{wrapfigure}
We assume that the candidate set could be large ($n=20\text{-}30$), but the final active set of non-zero merge coefficients will be small (see Sec.~\ref{sec:how}).
We propose splitting iterations into two stages: an initial stage that searches all merging coefficients $\alpha_i$ over $\mathcal{A}=\Delta_B$ to encourage sparse, meaningful combinations, followed by a final, polishing stage that fixes the sparsity pattern of $\balpha$ to admit only a few non-zero entries ($z<n$) and optimizes over $\mathcal{A}=[0,1]^z$ for unconstrained precision and exploration of this now limited space.
After $T_1$ initial-stage iterations, we extract the sparsity pattern of the current top-ranked $\balpha$, then proceed with $T_2$ final-stage iterations (we set $T_1=10, T_2=9$ for all experiments). 
The final-stage starts the entire process over, \emph{except} we retain any existing samples with same $z$ active coefficients.
The initial stage tends to converge to a moderately successful solution with stable sparsity pattern, then the final stage further boosts performance (inset).

\subsection{Content control}
So far, we have described a human-in-the-loop preferential Bayesian optimization that places no requirements on the image generator $g$.
However, style-content entanglement is a well-known challenge in image stylization, and fine-tuning-based methods are no exception: small parameter changes can unexpectedly alter content and confuse the user.
To address this, we apply lightweight content control using SDEdit~\cite{sdedit}, with a reference image generated by the base model (no adapters), preserving content at minimal extra cost so users can focus on style.
Compared to other methods~\cite{controlnet,paircustomization}, SDEdit better preserves style, achieving a style similarity~\cite{somepalli2024measuring} of 0.77 (1 for a perfect match) at control strength 0.2 in our default setting (see supplemental).

\subsection{Implementation Details}
We implemented the backend of \toolname in python using BoTorch. We defer to our supplemental for detailed hyperparameter values.

\begin{wrapfigure}{r}{0.35\linewidth}
\includegraphics[width=\linewidth,trim=0.95cm 0.8cm 0.0cm 0.9cm]{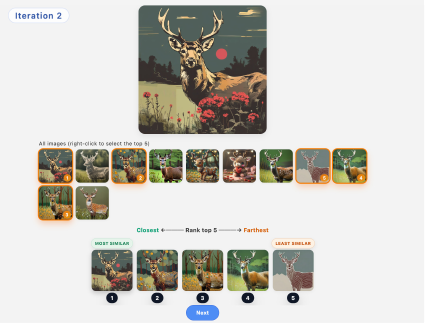}
\end{wrapfigure}
\paragraph{User Interface}
Our frontend user interface is a web application (see inset screenshot).
To reduce user fatigue, $N$ generated images are presented to the user pre-sorted by corresponding surrogate estimate, and the user is only asked to identify and sort the top $k \leq N$.
Whichever image is currently selected is shown in full resolution.

\section{Evaluations}
\label{sec:evaluations}

We evaluate our approach through quantitative experiments with both simulated and real users (Sec.~\ref{sec:quant}) and further demonstrate several extensions (Sec.~\ref{sec:qual}).

\subsection{Quantitative Experiments}
\label{sec:quant}

We evaluate using a \emph{matching task}, a standard protocol for assessing human-in-the-loop BO methods~\cite{brochu2010bayesian,koyama2017sequential,koyama2020sequential}.
The task measures how efficiently a method recovers a target solution under limited evaluation budgets, approximating practical scenarios where users have a clear goal, while enabling quantitative evaluation via known ground truth (GT).

Given an adapter collection of size $n$, test prompts $\{c^{(1)},\dots,c^{(m)}\}$, and corresponding GT coefficients $\{\balpha_{GT}^{(1)},\dots,\balpha_{GT}^{(m)}\}$, the task is
\begin{align*}
    \max_{\balpha \in \mathcal{A}} f(g(\balpha, c^{(i)}), \I_{GT}^{(i)}),
\end{align*}
where $\I_{GT}^{(i)} = g(\balpha_{GT}^{(i)}, c^{(i)})$ is not observed by our method and is available only to the simulated user.
We allow 5 initial inferences for initialization, followed by 20 optimization iterations with 8 inferences per iteration, reflecting a realistic inference and interaction budget for human-in-the-loop settings.

\begin{figure*}[t]{\textbf{}}
  \includegraphics[width=\linewidth]{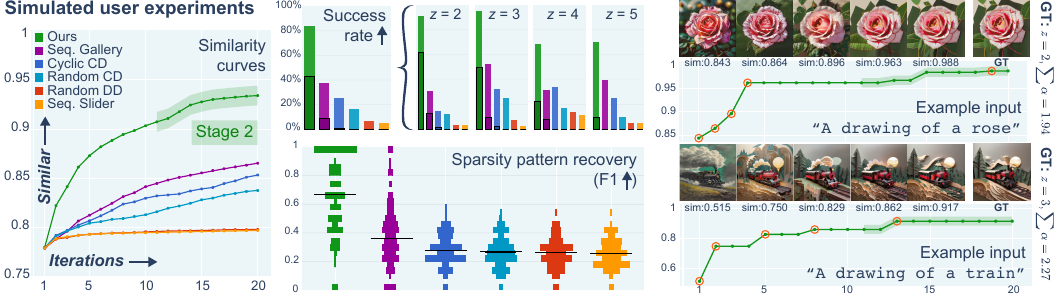}
  \caption{
  Running-best similarity curves averaged over 30 prompt-weight combinations and 5 random seeds, together with success rates and sparsity recovery (F1; black horizontal lines indicate the median) of the final best result.
  Success rates are first reported overall and then broken down by the ground-truth number of active adapters $z$, measured at a similarity threshold of $0.9$ (solid bars), with darker outlines indicating cases exceeding $0.95$.
  Curves are shown for two individual test inputs, with images corresponding to the highlighted iterations (orange circles).
  CD: coordinate descent; DD: directional descent.
  }
  \label{fig:evaluation}
\end{figure*}

\paragraph{Experiment setup}
We construct an adapter collection, a set of test prompts, and corresponding target coefficients.
The adapter collection is assembled from popular LoRA adapters on Civitai, manually selected to cover diverse styles and to produce images consistent with their advertised behavior under our local inference implementation.
While \toolname supports general fine-tuned models, including full-weight fine-tunes and other parameter-efficient adaptations, we use LoRAs due to their widespread adoption and straightforward integration.
Our primary simulated and real-user experiments use collections of size $n=20$.
To assess scalability, we additionally conduct simulated stress tests with $n=30$ and $40$.
We evaluate $m=30$ prompts (five runs per prompt with different random seeds) of the form ``a drawing of \texttt{<noun>}'' (or ``a portrait of \texttt{<noun>}'' for person nouns), with nouns drawn from CIFAR-10 and CIFAR-100 super-classes.
Target coefficients are sampled to match the empirical distribution observed in real usage statistics (Sec.~\ref{sec:how}); further details are provided in the supplemental.

\paragraph{Comparisons}
We compare against both line-search- and BO-based approaches.
For line search, we evaluate cyclic and random coordinate descent, mimicking the slider-based workflow where one coefficient is adjusted at a time, as well as random directional descent, which samples a random slider direction.  
Each slider spans the full coefficient hypercube, and line-search baselines use evenly spaced samples (including endpoints), as common line-search algorithms (e.g., Brent’s method) probe primarily near initial points under a limited budget of eight evaluations and often fail to escape local minima.
The two prior general BO methods rely on oracle local search via a slider~\cite{koyama2017sequential} and on a design gallery~\cite{koyama2020sequential}, which we set to use 3×3 grids (no zoom-ins) matching our budget of 8 inferences per iteration.
We note that the gallery approach comprises both an optimization scheme and a dedicated interface design, which can offer benefits such as reduced cognitive load through structured layouts and zoom-in capabilities for detailed inspection.
In our experiments, the zoom-in functionality is disabled to match a fixed, limited inference budget across conditions, which may diminish some of the advantages of the original interface design.
Therefore, the results should be interpreted as a comparison of optimization schemes under a controlled interaction budget, rather than a full system-level comparison.

\paragraph{Performance metrics}
We evaluate result quality using DreamSim~\cite{fu2023dreamsim}, a perception-aligned image similarity model.
Given an image output $g(\balpha)$ and a target image $\I_{GT}$, we define the objective as
$f(g(\balpha), \I_{GT}) = \text{DreamSim}(g(\balpha), \I_{GT})$, normalized to $[0,1]$ ($1$ indicates an exact match).
As shown in Fig.~\ref{fig:evaluation}, DreamSim aligns well with human judgments; empirically, scores of $0.9$ and $0.95$ correspond to rough and near-exact matching, respectively.
Additionally, we evaluate recovery of the GT active adapters using F1, which penalizes both missing and redundant adapters.
This metric complements image-based similarity by capturing whether high-quality results are achieved using the correct adapter subset.
Qualitative examples are in the supplemental material.

\subsubsection{Simulated user experiments}
We simulate human pairwise preference feedback using DreamSim, defining
$\balpha_i \succ \balpha_j$ whenever
$\text{DreamSim}(g(\balpha_i), \I_{GT}) > \text{DreamSim}(g(\balpha_j), \I_{GT})$.
The target image $\I_{GT}$ is only observed by the simulated user, \emph{not} by \toolname.

\begin{figure*}[t]{\textbf{}}
  \includegraphics[width=\linewidth]{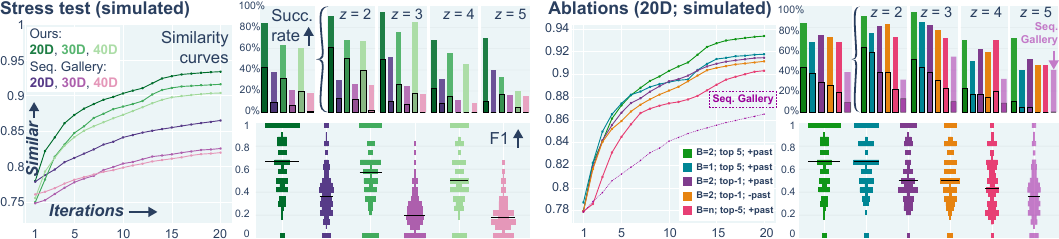}
  \caption{
    Left: running-best similarity curves under increasing dimensionality (30D and 40D), comparing \toolname to Sequential Gallery~\cite{koyama2020sequential}; \toolname degrades gracefully and consistently outperforms Sequential Gallery.
    Right: ablations in our default 20D setting, varying the simplex constraint $B \in \{1,2,n\}$ ($B=n$ means unconstrained), top-$k$ selection ($k=1$ vs.\ $5$), and inclusion of two past samples.
    The default configuration ($B=2$, top-$5$, $+$past) performs best.
    Importantly, even with a matched interaction budget ($B=2$, top-$1$, $-$past), \toolname substantially outperforms Sequential Gallery.
    }
  \label{fig:ablation}
\end{figure*}

We report running-best similarities across iterations, together with success rates and sparsity recovery (F1) of the final best result (Fig.~\ref{fig:evaluation}).
We compare three classes of methods:
\emph{Batch BO-based} includes our method ($q{=}8$) and Sequential Gallery~\cite{koyama2020sequential}, which can be viewed as a small-batch BO variant ($q{=}2$);
\emph{Coordinate-descent} updates one adapter at a time, including cyclic and random coordinate descent;
\emph{Direction-descent} includes random direction descent and the BO-powered Sequential Slider~\cite{koyama2017sequential}, which operates on a single sample per iteration.
Direction-descent methods struggle in high-dimensional spaces due to the low probability of sampling informative directions, while coordinate-descent methods achieve moderate gains by explicitly activating or deactivating individual adapters but struggle as dimensionality increases.
Batch BO-based methods achieve substantially better performance, demonstrating the benefits of informed exploration.
\toolname achieves rapid Stage~1 gains via the capped simplex constraint (Sec.~\ref{sec:simplex}), with further improvement at the onset of Stage~2.
As the GT number of active adapters $z$ increases, success rates decline, reflecting the greater difficulty of recovering more complex merges; yet \toolname consistently achieves the highest sparsity recovery, with many cases reaching $\text{F1}{=}1$.
Note that the two-stage strategy may miss the global optimum (i.e., $\text{F1} \neq 1$) if Stage 1 fails to include all true active adapters.
Finally, in simulated stress tests at higher dimensionalities (30D–40D), \toolname degrades gracefully and consistently outperforms the comparison (Fig.~\ref{fig:ablation}).

\begin{figure}[t]{\textbf{}}
  \includegraphics[width=\linewidth]{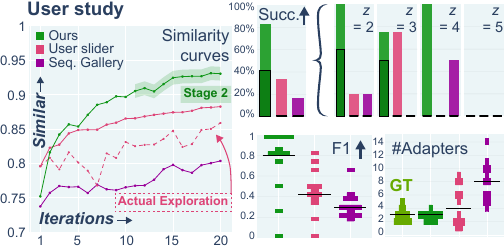}
  \caption{
    Running-best similarities averaged over 12 input-user combinations.
    For the slider interface, both running-best (solid) and explored intermediate results (dashed) are shown.
    We report success rates ($>0.9$ in solid and $>0.95$ outlined), sparsity recovery (F1; black lines indicate the median), and the number of active adapters, with ground truth indicated.
  }
  \label{fig:study}
\end{figure}

\subsubsection{User study}
\label{sec:study}
We conduct a user study to evaluate \toolname in a controlled human-in-the-loop setting.
Participants perform the same matching task using a subset of our 20D simulated experiment test inputs and the same adapter collection.
We compare three interaction interfaces: (1) a \emph{slider-based} interface commonly used in current adapter-merging workflows, (2) a \emph{gallery-based} interface implementing Sequential Gallery (\emph{gallery} for short), and (3) our proposed \emph{top-$k$ ranking} interface.
The slider interface serves as a more realistic baseline than coordinate-descent approaches, offering richer interactions (e.g., per-adapter examples, multi-slider updates, and history); it is initialized at $\balpha=\mathbf{0}$, with performance evaluated from the first user-modified result.
After a tutorial and practice session, participants complete all three interface conditions for 20 iterations, using different inputs and with order counterbalanced across participants (study details in the supplemental).

We recruit 12 participants (9 male and 3 female; 8 aged 25-34 and 4 aged 18-24) and record per-step user interactions, analyzed using the same DreamSim metrics (Fig.~\ref{fig:study}).
We report similarity curves averaged across users and target inputs.
For gallery and \toolname, curves reflect images explicitly selected by users, while for slider we report the running-best results (solid), which serve as an upper bound; the dashed curve shows the actual intermediate explorations and exhibits fluctuations, indicating difficulty in anticipating slider effects.
All methods show a statistically significant increase in similarity by the end of the session (gallery $p<0.05$, slider $p<0.05$, ours $p<10^{-3}$), with our method achieving the highest performance ($>0.9$ on average).
Consistent with simulation, our method achieves higher success rates and F1, accurately recovering the GT adapter set, while baselines activate redundant adapters.
Performance typically saturates by $\sim$15, suggesting the potential to reduce user fatigue with fewer iterations.
Accounting for user interaction time only, and excluding image generation latency ($\sim$10--15\,s per batch), the mean interaction time per step is 10.6\,s (gallery), 34.7\,s (slider), and 50.5\,s (ours).
To match user interaction time budgets, we take the total interaction time of 20 slider or gallery steps and truncate \toolname's trajectory to the number of steps that fit within the same budget, which is approximately 14 (slider) and 4 (gallery) steps.
Under this setup, our method outperforms slider with statistical significance ($p{<}0.05$) and achieves higher median similarity than gallery, though the difference is not statistically significant ($p{=}0.397$), as the matched budget corresponds to an early BO stage.
In practice, iteration time is often dominated by model inference rather than user interaction, motivating our focus on iteration budgets.
Including inference time in the comparison (15\,s per iteration in our setup), our method outperforms both baselines with statistical significance ($p{<}0.05$).
This advantage is expected to be larger in more realistic settings with limited compute resources (e.g., 1-2 local GPUs or online job queues), where inference can easily exceed one minute per iteration: with 60s inference, our method achieves higher median similarity (0.956 vs. 0.898 for slider; 0.926 vs. 0.854 for gallery), with statistical significance.
See Sec.~\ref{sec:future} for a further discussion about interaction time.

\begin{figure*}[t]
  \includegraphics[width=\linewidth]{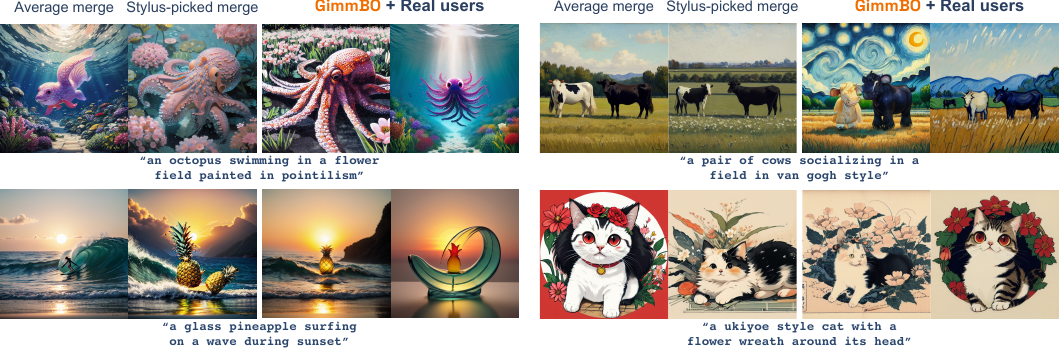}
  \caption{
    Given a text prompt, 20-25 community-shared adapters are gathered via coarse text-based retrieval of Stylus~\cite{luo2024stylus}
    Three methods are applied to determine the adapter merging configurations: trivial averaging all retrieved adapters, Stylus refinement, and \toolname.
    While averaging often introduces artifacts and Stylus-picked configurations yield a single fixed result, our method produces multiple, diverse, prompt-aligned outcomes per prompt through user-guided exploration (each pair starts from the same initialization and differs only due to user interactions).
    Examples are generated by SD1.5.
  }
  \label{fig:integrate}
\end{figure*}

\subsubsection{Ablations}
\label{sec:ablate}
We ablate using our simulated 20D matching task by varying three key factors: the simplex constraint parameter $B \in \{1,2,n\}$ ($n$ means simplex constraint off), top-$k$ selection ($k=1, 5$), and inclusion of two past samples ($+$past vs.\ $-$past) (Fig.~\ref{fig:ablation}, right).
Our default configuration ($B=2$, top-5, $+$past) performs best overall.
Specifically, the B-capped simplex constraint ($B<n$) improves performance, top-5 selection outperforms top-1 selection, and incorporating past samples provides additional gains.
Importantly, even with a matched interaction budget to Sequential Gallery ($B=2$, top-1, $-$past), \toolname substantially outperforms, with gains attributable to method design rather than increased user effort.

\begin{figure*}[p]
  \centering

  \begin{subfigure}{\textwidth}
    \centering
    \includegraphics[width=\textwidth,keepaspectratio]{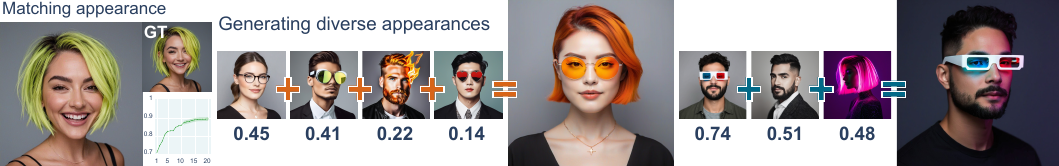}
    \caption{
      Appearance matching to a reference (left, GT) and diverse appearances generated by merging external appearance attributes such as hair, glasses, and expressions (right, indicated by various single-adapter generations), demonstrating generalization beyond style.
    }
    \label{fig:faces}
  \end{subfigure}

  \vspace{0.8em}

  \begin{subfigure}{\textwidth}
    \centering
    \includegraphics[width=\textwidth,keepaspectratio]{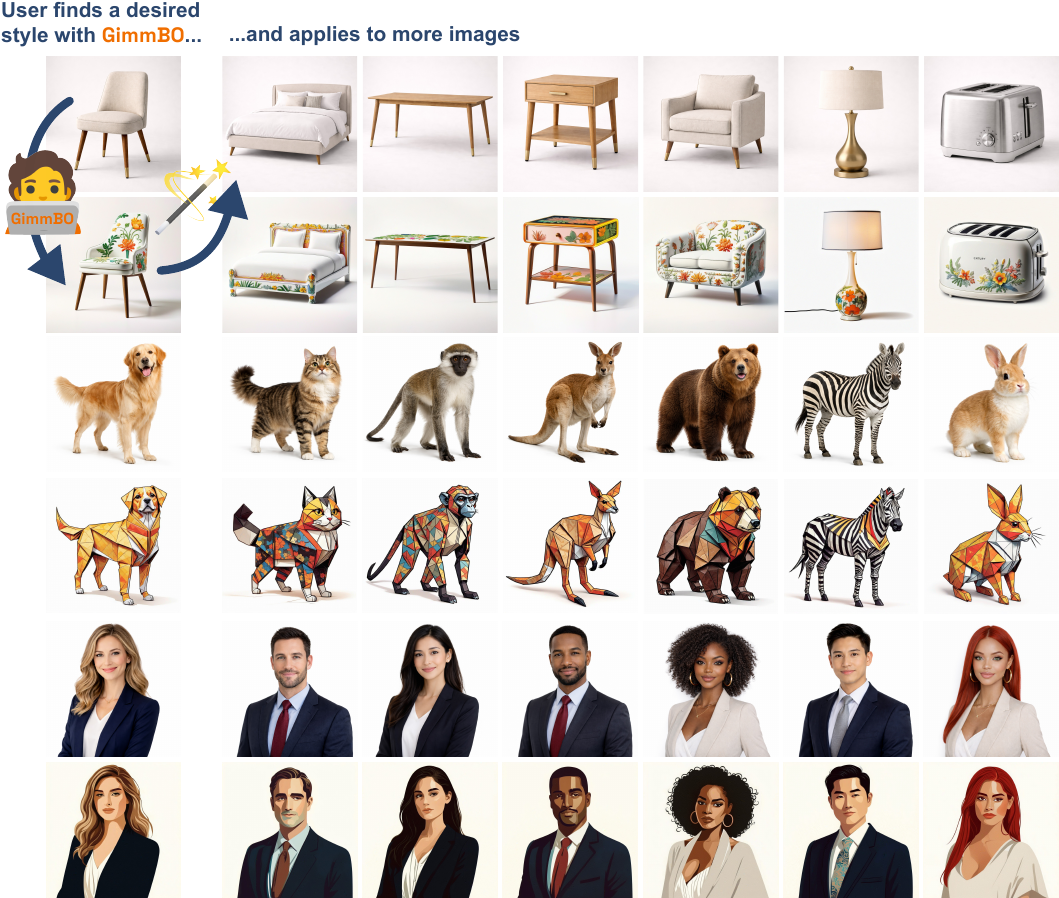}
    \caption{
      Applying a user-identified adapter merging configuration discovered with \toolname to new inputs via SDEdit achieves consistent stylization across images.
      All styles shown are discovered by real users; control images here are generated using ChatGPT, but can be arbitrary user-provided images.
    }
    \label{fig:content}
  \end{subfigure}

  \caption{
    Applications of adapter merging discovered with \toolname.
    Disclaimer: Image generators and style adapters may not preserve facial identities or perceived personal characteristics (e.g., race, gender); adapter merging configurations found via \toolname are no exception.
  }
  \label{fig:applications}
\end{figure*}

\subsection{Extensions}
\label{sec:qual}

We explore several extensions of \toolname.
All qualitative results demonstrating visual diversity start from the same initialization and differ only through user interactions (20 rounds).

\paragraph{Integration with coarse retrieval}
We integrate \toolname with Stylus's~\cite{luo2024stylus} coarse adapter retrieval (Fig.~\ref{fig:integrate}).
Given a text prompt, we configure Stylus to retrieve 25 candidate adapters via keyword retrieval, of which 19–23 remain after filtering.
We test prompts containing both content and style attributes.
Because Stylus only releases precomputed retrieval embeddings for Stable Diffusion~1.5 (SD1.5), we conduct this experiment on SD1.5, additionally demonstrating generalization beyond SDXL.
We compare against two baselines: uniformly averaging retrieved adapters and Stylus-refined coefficients.
Uniform averaging often introduces artifacts due to irrelevant or mis-retrieved adapters (e.g., goldfish for the octopus prompt), while Stylus refinement yields a single fixed solution.
In contrast, even in this challenging setting where SD1.5 is less expressive and community-shared adapters vary widely in quality, \toolname enables effective navigation of the retrieved adapter space, producing diverse, high-quality, and prompt-aligned results.

\paragraph{Content merging}
We explore assembling and merging external appearance attributes, including hair, glasses, and expressions, for face generation (Fig.~\ref{fig:faces}).
Using a 25D adapter set, we verify that our method generalizes from style to content through a small quantitative matching experiment on three faces with five seeds each, achieving a success rate of 67\%.
We further present qualitative results demonstrating diverse attribute combinations for two faces.

\paragraph{Downstream applications}
Beyond single image generation, an adapter merging configuration can be reused to consistently stylize new images.
In a downstream content-creation scenario, users identify a desired configuration with \toolname and apply it to generate related visual assets by using input images as SDEdit control inputs.

\section{Limitations and Future Work}
\label{sec:future}
We focus on linear model merging (Eq.~\ref{equ:merge}) for evaluation, though \toolname does not rely on linearity and can apply to other merging methods~\cite{matena2021,Yadav2023,Zheng2024}, left as future work.
It would be interesting to examine how often preference transitivity is violated in our setting (cf.~\cite{TverskyKahneman1981}) and how to address such cases if necessary.
The Dirichlet process used for acquisition sampling is fast and continuous, but this reparameterization introduces a known coordinate-ordering bias.
While alternatives such as uniform sampling the simplex~\cite{jacobson2023cappedsimplex} and projection-based methods~\cite{WangL15b} avoid this bias, our initial experiments suggest that it may be serendipitously beneficial in our setting.
Understanding this effect more systematically is left for future work.
DreamSim, used for our simulated user and performance metric, measures perceptual similarity rather than style specifically, and may conflate content and style.
We mitigate this with content control and complementary metrics, and future work could explore more style-specific evaluation measures.

\toolname with default hyperparameters incurs noticeably more user interaction time than the alternatives (Sec.~\ref{sec:study}).
Interaction effort can be reduced by adjusting interface parameters such as using a smaller top-$k$ and disabling past samples, which incurs only a small similarity drop while still outperforming baselines in matched-budget settings (Sec.~\ref{sec:ablate}).
Additional real-user studies to evaluate these trade-offs and explore lower-interaction paradigms are left for future work.
When users have limited access to compute (e.g., commercial cloud settings), per-iteration image generation will take significantly longer than user ranking.
This invites extensions of our work to hide latency with anticipatory system designs.

Beyond adapter merging, it remains an open question how our formulation extends to other increasingly popular customization regimes such as reference-image conditioning~\cite{ye2023ip,wang2024instantstyle,wang2025styleadapter} and style-injection pipelines~\cite{patashnik2025nested}.
These approaches offer alternative mechanisms for controlling style, often operating in latent or feature spaces rather than explicit parameter combinations.
It is unclear which aspects of our method, including preference-based search, sparsity structure, and staged exploration, would transfer directly.
Investigating such generalizations is a promising direction for future work.

Reinforcement learning from human feedback (RLHF) has been explored for improving generative models, including diffusion models, but typically relies on large-scale feedback datasets~\cite{yang2024using} or automatic reward models~\cite{miao2024subject}.
This differs from our setting: a single user with a limited interaction budget, where data efficiency is critical.
As a result, existing RLHF approaches are less directly applicable, though making RLHF viable in such low-feedback settings is an interesting direction for future work.

Generative image models continue to be used in unethical and harmful ways; adapter models contribute to and possibly amplify these uses.
Many of the adapters on Civitai target the sexualization of women via pose, clothing, or explicitly graphic subject matter. 
Using these models or \toolname mergers of them to manipulate photos of real people without their consent is clearly unethical. It is not clear how these problems are to be solved or alleviated with further technology research.

A recently released proprietary and closed-source application,
Style Creator\footnote{\url{https://docs.midjourney.com/hc/en-us/articles/41308374558221-Style-Creator}, released November 20, 2025}, offers related capabilities to \toolname, while differing in both interaction and underlying representation.
We hope that our work inspires further academic research on this problem.

\section{Acknowledgment}

We thank Yuki Koyama for clarifications on prior work, Leping Qiu for assistance with user study design, all anonymous participants in our pilot and final studies, and the DGP members at the University of Toronto for discussions and support.
Our research is funded in part by NSERC Discovery (RGPIN–2022–04680), the Ontario Early Research Award program, the Canada Research Chairs Program, a Sloan Research Fellowship, the DSI Catalyst Grant program, Arts \& Science Postdoctoral Fellowship at the University of Toronto, and gifts by Adobe Inc.

\bibliographystyle{ACM-Reference-Format}
\bibliography{main}

\appendix

\section{Implementation Details}

\toolname relies on a number of hyperparameters. We list their default or typical values/ranges in order of first appearance in our main text:
\begin{table}[h!]
\centering
\begin{tabular}{lll}
\hline
\textbf{Symbol} & \textbf{Description} & \textbf{Typical value} \\
\hline
$n$   & Size of candidate set                         & $[2,30]$ \\
$k$   & \# images ranked per iteration         & $5$ \\
$N$   & \# images shown per iteration          & $[5,10]$ \\
$B$   & Merge coefficient sum bound                   & $2$ \\
$\tau$ & Initialization sparsity threshold             & $0.1$ \\
$\lambda$ & UCB acquisition parameter                & $9$ \\
$q$   & \# batch samples                      & $\leq N$ \\
$S$   & \# L-BFGS-B seeds                     & $1024$ \\
$T_1, T_2$ & \# iterations per stage        & $10$ \\
\hline
\end{tabular}
\label{tab:parameters}
\end{table}

\subsection{Bayesian Optimization}
We consider adapter merging task with a 20–30D high-dimensional design space in which effective solutions are typically sparse.
We therefore require BO solvers that scale to high dimensions while identifying influential variables.
We considered SAASBO~\cite{eriksson2021high} and SEBO~\cite{liu2023sparse}, both of which employ a sparse axis-aligned subspace (SAAS) prior for automatic variable relevance detection, with SEBO further encouraging sparsity via an explicit $L_0$ regularization in a multi-objective formulation.
In preliminary experiments, we observed comparable performance between the two approaches; however, due to the additional memory and computational overhead introduced by SEBO’s homotopy-continuation-based optimization, we adopt the simpler SAASBO solver in our framework.

We implement the BO backend using \texttt{BoTorch}~\cite{balandat2020botorch}, with custom extensions for preference-based feedback and variable reparameterization.
The surrogate model is a Gaussian process equipped with a SAAS prior, enabling automatic identification of influential dimensions in high-dimensional design spaces.

SAAS-prior BO solvers are originally designed for scalar objectives, so we extend them to handle preference-based feedback.
While a fully Bayesian treatment naturally extends to preference learning by modeling pairwise comparisons, it becomes inefficient in practice, as the dimensionality grows while more samples are accumulated across iterations, increasing sampling cost and complicating acquisition optimization.
To address this issue, we adopt a hybrid strategy.
Given the user feedback represented by preference pairs
\[
\mathcal{D} = \left\{ (\balpha_i, \balpha_j) \;\middle|\; \balpha_i \succ \balpha_j \right\},
\]
we assume each configuration $\balpha\in\mathcal{A}$ is associated with an unobserved scalar \emph{latent utility} $u(\balpha)\in\mathbb{R}$, which reflects how well the corresponding outcome aligns with the user's subjective preference.
A reported preference $\balpha_i \succ \balpha_j$ indicates that $u(\balpha_i)$ is larger than $u(\balpha_j)$, up to noise in human judgment.
Accordingly, each comparison is modeled as a noisy utility difference
\[
u(\balpha_i) - u(\balpha_j) + \epsilon_{ij},
\qquad
\epsilon_{ij} \sim \mathcal{N}(0,\sigma^2),
\]
and the probability of observing the preference $\balpha_i \succ \balpha_j$ is modeled with pairwise probit-likelihood from preference learning
\[
p\!\left(\balpha_i \succ \balpha_j \mid u\right)
=
\Phi\!\left(\frac{u(\balpha_i)-u(\balpha_j)}{\sigma}\right),
\]
where $\Phi(\cdot)$ denotes the standard normal cumulative distribution function.
Latent utilities are estimated via maximum a posteriori (MAP) inference from comparisons using \texttt{BoTorch}'s default \texttt{PairwiseProbitLikelihood} implementation, and are then treated as scalar observations, while GP hyperparameters remain Bayesian under the SAAS prior.

Posterior inference over GP hyperparameters is performed using Hamiltonian Monte Carlo with the No-U-Turn Sampler (NUTS)~\cite{hoffman2014no}, implemented via \texttt{NumPyro} (a faster JAX backend) rather than the default \texttt{Pyro} implementation.
We use a single-chain NUTS sampler with $80$ warmup steps and $80$ posterior samples per iteration, a maximum tree depth of $6$, and median-based initialization.
To reduce correlation between samples, we apply thinning with a factor of $5$ when estimating acquisition functions.
This sampling procedure takes around $3$ seconds on average.

The resulting predictive posterior, represented as a Gaussian mixture, defines the acquisition function, which is optimized using gradient-based methods.
Batch candidate selection is performed via joint acquisition optimization with batch size $q=8$.
We use L-BFGS-B with $1024$ raw samples and $20$ random restarts to approximately solve the resulting global acquisition optimization problem.
Acquisition function optimization takes approximately $10$ seconds on average.
This step uses the standard \texttt{SciPy}~\cite{2020SciPy-NMeth} implementation of L-BFGS-B, which is not performance-optimized for this setting and represents a potential source of further speedup.

Due to the stick-breaking mapping from $\x \in [0,1]^n$ to the capped simplex, gradients with respect to $\x$ can vanish near $x_i \in {0,1}$.
In practice, we observe premature termination of gradient-based acquisition optimization due to this effect.
To address this issue, we clamp variables to a slightly contracted interval $[10^{-4}, 1-10^{-4}]$ during acquisition optimization, and subsequently round values to $0$ or $1$ using a threshold of $10^{-2}$.
As a result, when extracting the sparsity pattern before the final stage we can test for exact non-zeros.

\subsection{Past Sampling}
At each iteration, we include several previously evaluated samples when presenting candidates to the user, which improves the accuracy and stability of latent utility estimation.
Because these samples have already been generated, they incur no additional inference cost.
The design of this mechanism has two considerations: (1) prioritizing samples with high estimated utility, and (2) avoiding repeated presentation of the same sample across iterations.
Let $\{y_i\}_{i=1}^M$ ($M$ indicating the number of accumulated samples) denote the current latent utility estimates inferred from preference feedback, after excluding the sample corresponding to the user’s current top-1 selection, which is always included by default.
We design a sampling procedure that selects up to $M_{\text{past}}$ past samples while explicitly accounting for both factors.

We first randomly shuffle these candidates.
We then convert latent utilities into sampling probabilities via a temperature-scaled softmax.
Latent utilities are first normalized to $[0,1]$ using min-max normalization; if all utilities are equal, the normalized utilities are set to zero to avoid numerical instability.
The normalized utilities are scaled by a temperature parameter $\sigma=2$, which controls the sharpness of the resulting distribution.

To discourage repeatedly selecting samples that already dominate the comparison history, we further adjust the normalized utilities using the existing comparison pairs.
For each sample, we count how many times it appears in the set of recorded comparison pairs and normalize these counts to sum to one.
This normalized comparison count is subtracted from the scaled utility, yielding an adjusted utility
\[
\tilde{u}(\balpha_i) = \frac{u(\balpha_i)}{\sigma} - C(\balpha_i),
\]
where $u_i$ denotes the normalized latent utility and $C(\balpha_i)$ the normalized comparison count.
Sampling probabilities are obtained by applying a softmax over $\{\tilde{u}(\balpha_i)\}$.
We then randomly draw from $M_{\text{past}}$ past candidates from $\{y_i\}_{i=1}^M$ excluding the current top-ranked sample according to this distribution.

\subsection{Image Generation}

All image generation is performed using an SDXL-based pipeline implemented in \texttt{ComfyUI}.
We use the DPM++~2M sampler~\cite{kumar2025dpm} with Karras et al.'s noise scheduler~\cite{karras2022elucidating}.
Adapter merging is implemented via ComfyUI’s LoRA stack mechanism, with weight interpretation set to \texttt{comfy} and token normalization disabled.
Unless otherwise specified, all experiments use a fixed inference configuration to isolate the effect of adapter weights.

Specifically, we use classifier-free guidance with a guidance scale of $7$.
The number of diffusion steps is fixed across experiments (\texttt{steps} = 30), and the same random seed is used for all methods to eliminate stochastic variation unrelated to adapter weights.
Images are generated at a fixed resolution of $1024\times1024$.
Inference batch size is set to $1$ throughout.

All prompts follow the template
\texttt{a drawing of <noun>}.
(or \texttt{a portrait of <noun>} when the noun refers to a person),
with nouns drawn from CIFAR-10 and CIFAR-100 super-classes as described in Sec.~\ref{sec:prompt}.
The same prompt is used across all compared methods for a given test case.

Prompts are augmented with a fixed negative prompt to suppress common artifacts, following standard community practice.
The full negative prompt used in all experiments is:
\begin{quote}\small
\texttt{easynegative, 3d, realistic, badhandv4, (lowres, bad quality), (worst quality, low quality:1.3), blurry, cropped, out of frame, border, bad hands, interlocked fingers, mutated hands, (bad anatomy:1.4), from afar, warped, (deformed, disfigured:1.1), twisted torso, mutated limbs}
\end{quote}

All inference hyperparameters, prompts, samplers, schedulers, and seeds are held constant across methods and experimental conditions.
Only the adapter weights produced by the optimization framework vary between runs.
This ensures that all observed differences are attributable solely to the adapter merging strategy rather than inference variability.

All experiments are run on NVIDIA \texttt{RTX A6000} and \texttt{RTX 6000 Ada} GPUs.
To reduce interaction latency, image inference is parallelized across multiple GPUs.
In our setup, generating a batch of $8$ images takes approximately $20$ seconds when distributed over $4$ GPUs in the simulated-user setting, and approximately $10$ seconds when distributed over $8$ GPUs in the real-user setting.

\subsection{Content Control}
\begin{figure}[t]
  \includegraphics[width=\linewidth]{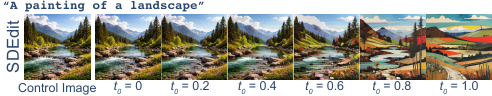}
  \caption{
    Given a text prompt, SDEdit~\cite{sdedit} initializes the denoising process from a provided control image and injects noise corresponding to timestep $t_0 \leq 1$.
    Smaller $t_0$ values preserve the control image more strongly (stronger control), while larger $t_0$ values allow greater deviation (weaker control).
  }
  \label{fig:control}
\end{figure}

Style-content disentanglement is a long-standing challenge in text-to-image generation, which can interfere with users' experiences with our proposed interface. We address this by applying SDEdit~\cite{sdedit} with $t_0 = 0.8$ across our experiments with simulated users and real users to achieve the optimal amount of content control without sacrificing stylization.
Essentially, SDEdit \cite{sdedit} initializes the denoising process with the provided control image and modifies the denoising process with noise initialized with timestep $t_0 \leq 1$. 
In the main text, we therefore report $t_0 = 0.8$ as a control strength of $0.2$ to reflect the inverse relationship between $t_0$ and the influence of the control image.
Empirically, we found SDEdit to achieve the best trade-off between stylization and content control among many other proposed techniques like ControlNet \cite{controlnet, paircustomization}. 
See Fig.~\ref{fig:control} for the generated image with and without the content control. 

In particular, we quantitatively evaluate the choice of $t_0 \in {0.6, 0.7, 0.8}$ when applying SDEdit in our generation process by measuring the style similarity score \cite{somepalli2024measuring} between the image generated with and without SDEdit averaged across 100 different combinations of style adapters and observed that 0.8 achieves the best trade-off ($t_0 \in \{0.6, 0.7, 0.8\}$ yields style similarity of 0.56, 0.70, and 0.77, respectively).

All simulated-user and real-user experiments, as well as the content-merging extension, are generated using SDEdit with a $t_0=0.8$.
The coarse-retrieval integration is produced without SDEdit control, as control images automatically generated by SD1.5 without adapters are generally low quality.
For consistent stylization across multiple images, we use $t_0=0.7$, which applies stronger content control than our default setting to better preserve the input images.

\section{Test Construction}
\label{sec:prompt}

\paragraph{Simulated user experiments}
We evaluate $m=30$ prompts with nouns drawn from CIFAR-10 and CIFAR-100 super-classes.
Prompts consist of ``a drawing of'' objects and animals
(bird, cat, deer, dog, frog, horse, dolphin, shark, butterfly, tiger,
elephant, turtle, airplane, car, ship, truck, rose, bottle, apple, lamp,
chair, house, mountain, oak, train)
and ``a portrait of'' people (baby, girl, boy, woman, man).
We uniformly sample merging coefficients for $n \in \{20,30,40\}$ and discard entries with magnitude below $0.1$.
We then construct $m=30$ merging coefficient vectors whose distributions of active adapter count and coefficient magnitude match the surveyed real-world statistics.
Samples are binned for statistical reporting in Table~\ref{tab:test_inputs}.
To avoid ground truth targets with low quality, we manually inspect and reduce magnitudes while preserving each sample’s bin assignment if necessary.

\paragraph{User study}
We randomly select 12 test inputs from our $m=30$ prompt-adapter combination set while preserving the same sparsity and magnitude distributions.
We assign the 12 test inputs to main study sessions such that each input is associated with all three interfaces, with each input-interface combination completed by a different user.
Participants perform the same matching task using a subset of prompts and adapter collections, with the ground-truth number of active adapters $z \in \{2,3,4,5\}$ (5, 4, 2, and 1 cases, respectively).
Samples are binned for statistical reporting in Table~\ref{tab:study_inputs}.

\begin{table}[h!]
\centering
\caption{Distribution of test input merging coefficients.}
\label{tab:test_inputs}
\footnotesize
\begin{tabular*}{0.8\linewidth}{@{\extracolsep{\fill}} c c|c c c c}
\noalign{\hrule height 1pt}
$z$ & \#Cases & \multicolumn{4}{c}{Weight sum range (\#Cases)} \\
           &         & $(0,1]$ & $(1,2]$ & $(2,3]$ & $(3,4]$ \\
\hline
2 & 11 & 1  & 10 & -- & -- \\
3 & 8  & -- & 3  & 5  & -- \\
4 & 7  & -- & 2  & 4  & 1  \\
5 & 4  & -- & 1  & 2  & 1  \\
\noalign{\hrule height 1pt}
\end{tabular*}
\end{table}

\begin{table}[h!]
\centering
\caption{Distribution of user study input merging coefficients.}
\label{tab:study_inputs}
\footnotesize
\begin{tabular*}{0.8\linewidth}{@{\extracolsep{\fill}} c c|c c c c}
\noalign{\hrule height 1pt}
$z$ & \#Cases & \multicolumn{4}{c}{Weight sum range (\#Cases)} \\
    &         & $(0,1]$ & $(1,2]$ & $(2,3]$ & $(3,4]$ \\
\hline
2 & 5 & -- & 5 & -- & -- \\
3 & 4 & -- & 2 & 2  & -- \\
4 & 2 & -- & -- & 2  & -- \\
5 & 1 & -- & -- & 1  & -- \\
\noalign{\hrule height 1pt}
\end{tabular*}
\end{table}

\subsection{User Study Design}

Our user study investigates how different optimization backends, as exposed through their corresponding interfaces, support human influence over image generation via preference-based feedback, with a particular focus on how effectively users can reach satisfactory results under limited computational and interaction budgets.
We refined the task description and interface designs with the help of feedback from four pilot study participants.

\begin{figure}[t]
  \includegraphics[width=\linewidth]{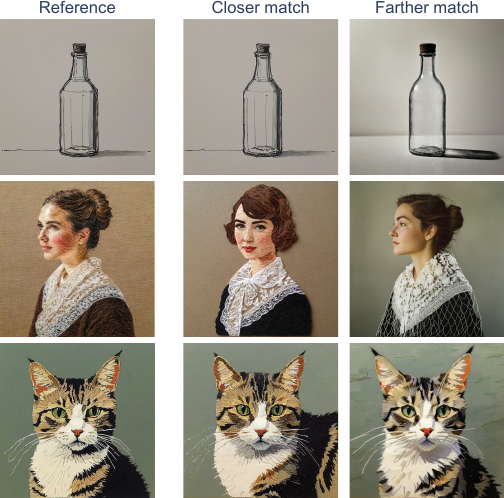}
  \caption{
    Example images displayed in our matching task description.
    These three adapter merging configurations never appear in the practice or main study sessions.
  }
  \label{fig:task}
\end{figure}

\begin{figure*}[t]{\textbf{}}
  \includegraphics[width=\linewidth]{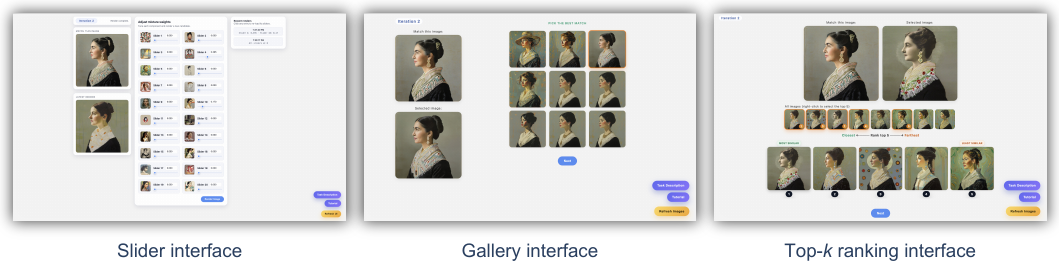}
  \caption{
    Interfaces shown to participants in the user study.
    Each interface displays the target image for the matching task and provides utility buttons (bottom right) for quick access to the task description and interface instructions any time during the study.
    This input is an example for our interface instructions, not used for the practice or actual sessions.
    }
  \label{fig:uis}
\end{figure*}

\subsection{Task Description}

Participants were presented with the following instructions at the start of the study:

\begin{quote}
In this study, we will ask you to compare a \textbf{reference image} against several \textbf{candidate images}.
Your task is to judge how visually similar each candidate is to the reference, and provide feedback based on your judgment.
Please focus on overall visual appearance and style, rather than image content (e.g., pose, composition, subject structure) or visual errors (e.g., incorrect anatomy).
Please do not base your decision on personal preference or taste.
The following examples illustrate what it means for an image to appear closer to or farther from a reference.
Please take your time and look closely at each example before proceeding.
\end{quote}

Example images illustrating closer and farther matches are shown in Fig.~\ref{fig:task}.
These examples are not part of the test input set.
All images can be displayed at their original resolution to allow careful visual inspection.

\subsection{User Interfaces}

All interfaces share a common task structure and a set of core UI components to ensure consistency across conditions (Fig.~\ref{fig:uis}).
Each interface displays an iteration counter indicating the user’s current progress.
A fixed reference image is shown together with one or more candidate images for comparison.
Depending on the interface, the reference and candidates are arranged either vertically or horizontally; in all cases, both can be expanded and viewed at their original resolution.
The interface provides persistent help controls, including access to the task description, an interface-specific tutorial, and an image refresh button.
All generated images are passed through an automatic safety filter based on the Stable Diffusion safety checker from CompVis\footnote{\url{https://huggingface.co/CompVis/stable-diffusion-safety-checker}}.
Images flagged as unsafe are blurred when displayed.

\paragraph{Slider-based interface.}
The slider-based interface exposes one continuous slider per adapter, allowing users to manually adjust mixture weights and render a new image on demand.
Sliders are initialized to zero at the beginning of each iteration and can be adjusted either by dragging or by directly entering numeric values.
To support revisiting previous configurations, the interface maintains a history of recent renders; selecting a history entry restores the corresponding slider values.
Only a single candidate image is generated and evaluated per iteration.

\paragraph{Gallery-based interface.}
The gallery-based interface presents a fixed grid of candidate images generated from different parameter settings within the current iteration.
Users select the single image they judge to be most visually similar to the reference image.
No explicit parameter controls are exposed; the interface functions as a discrete selection mechanism over generated candidates.
After a selection is made, the system proceeds to the next iteration.

\paragraph{Top-$k$ ranking interface.}
Our proposed interface presents a set of candidate images generated at each iteration.
Users are asked to select and rank only the top-$k$ images according to their visual similarity to the reference, from most similar to least similar.
Candidates are pre-ordered by the current surrogate utility estimate to reduce interaction overhead.
To improve stability across iterations, a small number of previously generated samples may be included alongside newly generated candidates.
After submitting a ranking, the interface advances to the next iteration.
The transition between Stage~1 and Stage~2 is fully transparent to users: after 11 interaction rounds (sorting the initial samples and 10 batches proposed by \toolname), optimization restarts for Stage~2 based on the current top-ranked sample, retaining Stage~1 samples with the same active adapters and initializing five new samples in this reduced search space.

\subsection{Demographic Details}

We surveyed 12 participants (9 male and 3 female; 8 aged 25--34 and 4 aged 18--24).
All participants had backgrounds in computer science, machine learning, or AI, with 4 also having prior experience in visual arts or design.
Participants reported varying familiarity with generative image tools: 7 moderately familiar, 3 slightly familiar, 1 very familiar, and 1 not at all familiar.
While not exhaustive, this participant pool broadly reflects common users of contemporary image-generation tools in terms of technical background and tool familiarity.
The user study was conducted in person, with participants performing the interface tasks through a web browser.
See Table~\ref{tab:participant_demographics} for additional details.

\begin{table}[h!]
\centering
\caption{Participant demographics and background (counts).}
\label{tab:participant_demographics}
\footnotesize
\setlength{\tabcolsep}{6pt}
\begin{tabular}{l r}
\hline
\textbf{Category} & \textbf{Count} \\
\hline

\multicolumn{2}{l}{\textbf{Age group}} \\
\hline
18--24 & 4 \\
25--34 & 8 \\
\hline

\multicolumn{2}{l}{\textbf{Gender identity}} \\
\hline
Man & 9 \\
Woman & 3 \\
\hline

\multicolumn{2}{l}{\textbf{Familiarity with generative models}} \\
\hline
Not at all familiar & 1 \\
Slightly familiar & 3 \\
Moderately familiar & 7 \\
Very familiar & 1 \\
\hline

\multicolumn{2}{l}{\textbf{Usage frequency}} \\
\hline
Never & 1 \\
Less than once a month & 9 \\
A few times a month & 1 \\
A few times a week & 1 \\
\hline

\multicolumn{2}{l}{\textbf{Experience depth}} \\
\hline
Only tried a few times & 7 \\
Personal or creative projects & 6 \\
Professional or academic work & 1 \\
Fine-tuned or customized models & 2 \\
\hline

\multicolumn{2}{l}{\textbf{Domain background} (multi-select)} \\
\hline
Computer science or engineering & 12 \\
Machine learning or AI & 12 \\
Visual arts or design & 4 \\
\hline
\end{tabular}
\end{table}

\paragraph{Post-study feedback}
We collected optional post-study questionnaires in which participants freely expressed their impressions of each interface.
For the slider-based interface, participants reported difficulty in both identifying relevant style components and adjusting their proportions, noting that ``it was really difficult to figure out the style components unless you try all the options'' and that refining proportions was ``also difficult,'' highlighting the challenge of combining discovery and weight tuning in a single workflow.
For the gallery-based interface, participants frequently reported getting ``stuck in a local minimum,'' particularly when many candidates appeared visually similar, which limited effective exploration.
In contrast, feedback on our interface emphasized usability and engagement: participants described it as ``fun to rank 5 options'' and appreciated that it ``broke up the process into digestible steps,'' suggesting that ranking-based interaction helps structure exploration and reduce cognitive load.
Participants also noted limitations for our interface, including the lack of explicit history editing (``I wanted to go back and redo some of the earlier choices'') and occasional difficulty discerning which choices were most meaningful when many options were presented.
These comments motivate future improvements to the interface by exposing additional control to users.
Potential directions include making sample histories visible and editable, allowing users to control the timing of the transition between stages, and enabling a non-fixed number of samples for top-$k$ ranking.

\end{document}